\documentclass[letterpaper]{article} 
\usepackage{aaai24}  
\usepackage{times}  
\usepackage{helvet}  
\usepackage{courier}  
\usepackage[hyphens]{url}  
\usepackage{graphicx} 
\usepackage{natbib}  
\usepackage{caption} 
\usepackage{algorithm}
\usepackage{algorithmic}

\usepackage{times}
\usepackage{epsfig}
\usepackage{graphicx}
\usepackage{amsmath}
\usepackage{amssymb}

\usepackage{booktabs}
\usepackage{multirow}
\usepackage{makecell}
\usepackage{enumitem}
\usepackage{colortbl}
\usepackage{tabulary}
\usepackage[table]{xcolor}

\usepackage{color} 
\usepackage{bbding}
\usepackage{subfigure}  
\usepackage{pifont}

\usepackage{newfloat}
\usepackage{listings}
\DeclareCaptionStyle{ruled}{labelfont=normalfont,labelsep=colon,strut=off} 
\floatstyle{ruled}
\newfloat{listing}{tb}{lst}{}
\floatname{listing}{Listing}
\title{Efficient Deweather Mixture-of-Experts with Uncertainty-aware \\ Feature-wise Linear Modulation}
\author{
    Rongyu Zhang\textsuperscript{\rm 1,2},
    Yulin Luo\textsuperscript{\rm 2},
    Jiaming Liu\textsuperscript{\rm 2},
    Huanrui Yang\textsuperscript{\rm 3},
    Zhen Dong\textsuperscript{\rm 3},
    Denis Gudovskiy\textsuperscript{\rm 4},
    Tomoyuki Okuno\textsuperscript{\rm 4},
    Yohei Nakata\textsuperscript{\rm 4},
    Kurt Keutzer\textsuperscript{\rm 3},
    Yuan Du\textsuperscript{\rm 1}\equalcontrib,
    Shanghang Zhang\textsuperscript{\rm 2}\equalcontrib
}
\affiliations{
    \textsuperscript{\rm 1}Nanjing University
    \textsuperscript{\rm 2}National Key Laboratory for Multimedia Information Processing, School of Computer Science, \\ Peking University
    \textsuperscript{\rm 3}University of California, Berkeley
    \textsuperscript{\rm 4}Panasonic

    yuandu@nju.edu.cn, shanghang@pku.edu.cn
}

\usepackage{bibentry}

\begin{document}

\maketitle

\begin{abstract}
The Mixture-of-Experts (MoE) approach has demonstrated outstanding scalability in multi-task learning including low-level upstream tasks such as concurrent removal of multiple adverse weather effects. 
However, the conventional MoE architecture with parallel Feed Forward Network (FFN) experts leads to significant parameter and computational overheads that hinder its efficient deployment. In addition, the na\"ive  MoE linear router is suboptimal in assigning task-specific features to multiple experts which limits its further scalability.  
In this work, we propose an efficient MoE architecture with weight sharing across the experts. Inspired by the idea of linear feature modulation (FM), our architecture implicitly instantiates multiple experts via learnable activation modulations on a single shared expert block. 
The proposed Feature Modulated Expert (\texttt{FME}) serves as a building block for the novel Mixture-of-Feature-Modulation-Experts (\texttt{MoFME}) architecture, which can scale up the number of experts with low overhead.
We further propose an Uncertainty-aware Router (\texttt{UaR}) to assign task-specific features to different FM modules with well-calibrated weights. This enables MoFME to effectively learn diverse expert functions for multiple tasks. 
The conducted experiments on the multi-deweather task show that our \texttt{MoFME} outperforms the baselines in the image restoration quality by 0.1-0.2 dB and achieves SOTA-compatible performance while saving more than 72\% of parameters and 39\% inference time over the conventional MoE counterpart. Experiments on the downstream segmentation and classification tasks further demonstrate the generalizability of \texttt{MoFME} to real open-world applications.
\end{abstract}

\section{Introduction}
There is a growing interest in low-level upstream tasks such as adverse weather removal (deweather)~\cite{valanarasu2022transweather}. It intends to eliminate the impact of weather-induced noise on decision-critical downstream tasks such as detection and segmentation~\cite{zamir2022restormer}. 
Previous methods~\cite{ren2019progressive, chen2021all} approach each type of weather effect independently, yet multiple effects can appear simultaneously in the real world. Moreover, such methods mainly focus on the deweathering performance metrics rather than an efficient deployment.

One promising way to address several weather effects concurrently is the conditional computation paradigm~\cite{bengio2013deep}, where a model can selectively activate certain parts of architecture, i.e. the task-specific experts, depending on the input. In particular, the sparse Mixture-of-Experts (MoE)~\cite{riquelme2021scaling} with parallel Feed Forward Network (FFN) experts rely on a router to activate a subset of FFNs for each weather-specific input image. 
Figure~\ref{fig:1} shows a pipeline with an upstream MoE model to overcome a number of weather effects. 
For example,~\citet{ye2022towards} propose the DAN-Net method that estimates gated attention maps for inputs and uses them to properly dispatch images to task-specific experts. Similarly,~\citet{luo2023mowe} develop a weather-aware router to assign an input image to a relevant expert without a weather-type label at test time.

\begin{figure}[t]
\centering
\subfigure[MoE]{\includegraphics[width=2.66cm]{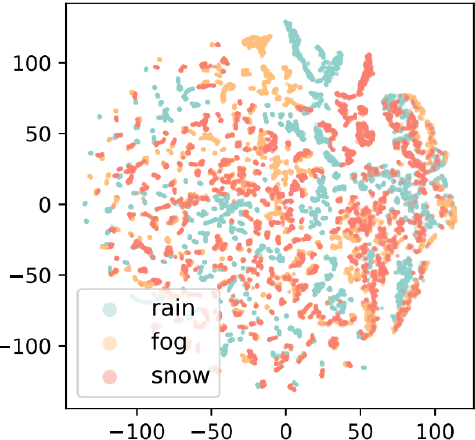}}
\subfigure[M$^{3}$Vit]{\includegraphics[width=2.7cm]{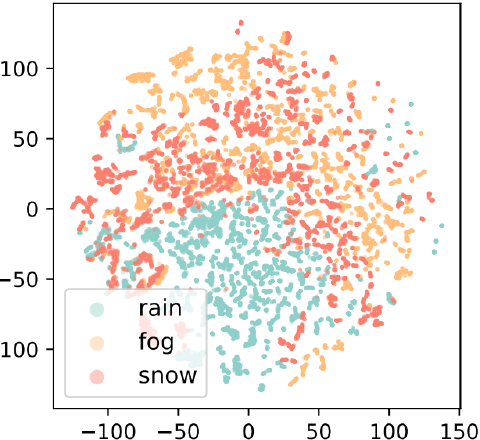}}
\subfigure[MoFME]{\includegraphics[width=2.6cm]{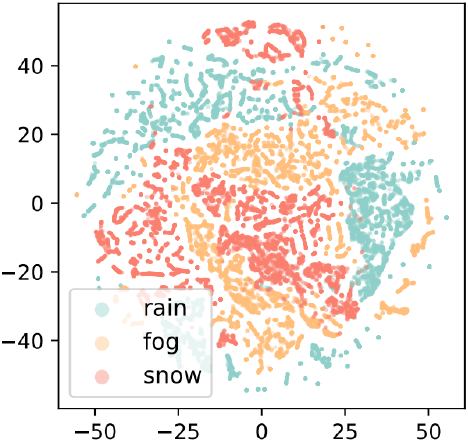}}
\vspace{-0.4cm}
\caption{t-SNE visualization of the router's outputs between different MoE architectures with adverse weather inputs.} 
\vspace{-0.6cm}
\label{fig:tsne} 
\end{figure}

Meanwhile, challenges exist in building a practical MoE-based model for deweather applications: \ding{202} \textit{Efficient deployment}. Conventional MoE-based models with multiple parallel FFN experts require a significant amount of memory and computing. For example, MoWE~\cite{luo2023mowe} architecture contains up to hundreds of experts with billions of parameters. Hence, it is infeasible to apply such architectures to edge devices with limited resources for practical upstream tasks, e.g. to increase the safety of autonomous driving\cite{chi2023bev}. Previous attempts to reduce memory and computation overheads inevitably sacrifice model performance~\cite{xue2022one}. 
\ding{203} \textit{Diverse feature calibration}. Existing MoE networks typically use na\"ive linear routers for expert selection. This leads to poor calibration of router weights with diverse input features. 
Multi-gate MoE~\cite{ma2018modeling} overcomes this challenge by designing an additional gating network to distinguish task-specific features. However, this introduces additional computation costs.
Therefore, we are motivated by the following objective: \textit{is it possible to design a computationally-efficient MoE model while improving its deweathering metrics for real-world applications?}

To approach this objective, we start by analyzing redundancies in the conventional MoE architecture.
The main one comes from multiple parallel experts containing independently learned weights. 
Meanwhile, previous research shows a possibility to simultaneously learn multiple objectives with diverse features using a mostly shared architecture and weights. 
For example, feature modulation (FM)~\cite{perez2018film,liu2021overfitting,liu2023vida} performs an input-dependent affine transformation of intermediate features with only two additional feature map parameters. Hence, the FM method allows decoupling multiple tasks simultaneously and implicitly represents ensemble models~\cite{turkoglu2022film} with low parameter overhead. 
Inspired by the FM method, we develop an efficient MoE architecture with feature-wise linear modulation for open-world scenarios. In particular, we propose \textbf{Mixture-of-Feature-Modulation-Experts} (MoFME) framework with two novel components: \textbf{Feature Modulated Expert} and \textbf{Uncertainty-aware Router}.

\textbf{FME} adopts FM into the MoE network via a single shared expert block. This block learns a diverse set of activation modulations with a minor overhead on the weight count. In particular, FME performs a feature-wise affine transformation on the model’s intermediate features that is conditioned on the task-specific inputs. 
Next, it fuses task-specific modulated features with a single shared FFN expert, which allows it to efficiently learn a set of input-conditioned models. Thus, FME increases generalization to a wider range of substantially different tasks during training. As the T-SNE visualization shown in Figure \ref{fig:tsne}, MoFME can better correlate the features with clearer partitions and boundaries.

The conventional MoE router adopts the top-$K$ mechanism, which introduces non-differentiable operations into the computational graph and complicates the router optimization process. Previous research has found that such MoE router is prone to mode collapse, where it tends to direct all inputs to a limited number of experts~\cite{riquelme2021scaling}. 
At the same time,~\citet{kendall2018multi} shows that uncertainty captures the relative confidence between tasks in the multi-task setting. Therefore, we propose our \textbf{UaR} router that estimates uncertainty using MC dropout~\cite{gal2016dropout}. The estimated uncertainty is used to weigh modulated features and, therefore, route them to the relevant experts.

We verify the proposed MoFME method by conducting experiments on the deweather task. For instance, evaluation results with All-weather~\cite{valanarasu2022transweather} and RainCityscapes~\cite{hu2019depth} datasets show that the proposed MoFME outperforms prior MoE-based model in the image restoration quality with less than 30\% of network parameters. In addition, quantitative results on the downstream segmentation and classification tasks after applying the proposed MoFME further demonstrate the benefits of our pipeline with the upstream pre-processing.
Our main contributions are summarized as:
\begin{itemize}
\setlength\itemsep{0em}
    \item We introduce Mixture-of-Feature-Modulation-Experts (MoFME) framework with two novel components to improve upstream deweathering performance while saving a significant number of parameters.
    
    \item We develop Feature Modulation Expert (FME), a novel MoE layer to replace the standard FFN layers, which leads to improved performance and parameter efficiency.

    \item We devise an Uncertainty-aware Router (UaR) to enhance the assignment of task-specific inputs to the subset of experts in our multi-task deweathering setting.

    \item Experimental results demonstrate that the proposed MoFME can achieve consistent performance gains on both low-level upstream and high-level downstream tasks: our method achieves \textbf{0.1-0.2 dB} PSNR gain in image restoration compared to prior MoE-based model and outperforms SOTA baselines in segmentation and classification tasks while saving more than \textbf{72\%} parameters and \textbf{39\%} inference time. 

\end{itemize}

\begin{figure*}[t]
\includegraphics[width=0.95\textwidth]{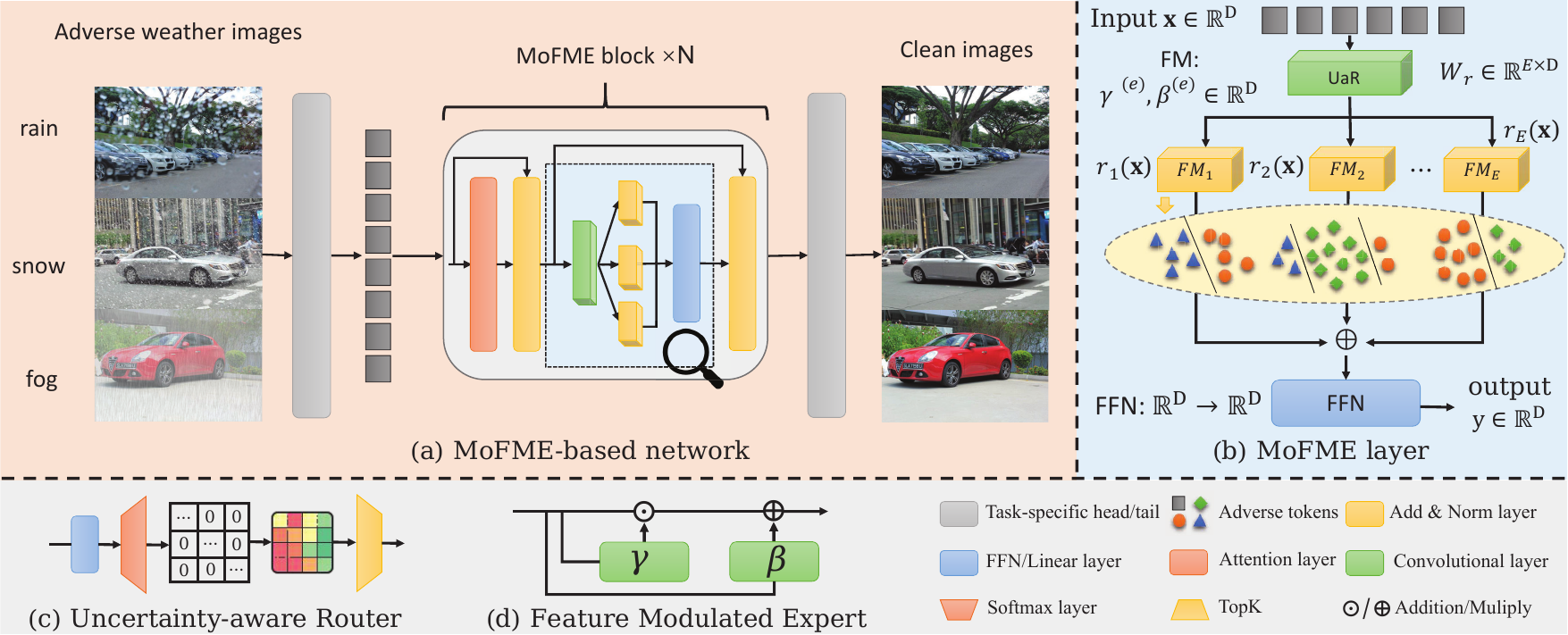}
\centering
\caption{Schematic illustration of the proposed (a) Mixture-of-Feature-Modulation-Experts (MoFME) network, and the (b) detailed MoFME layer with two novel components (c) Uncertainty-aware Router and (d) Feature Modulated Expert.}
\label{fig:method}
\vspace{-0.2cm}
\end{figure*}

\section{Related Work}
\noindent\textbf{Mixture-of-Experts (MoE).}
Sub-model assembling is a typical way to scale up model size and improve performance in deep learning. MoE is a special case of assembling with a series of sub-models that are called the experts. It performs conditional computation using an input-dependent scheme to improve sub-model efficiency~\cite{sener2018multi,jacobs1991adaptive,jordan1994hierarchical}. 
Specifically,~\citet{eigen2013learning,ma2018modeling} assemble mixture-of-experts models into an architectural block that is known as the MoE layer. This enables more expressive modeling and decreases computation costs. 
Another solution is to sparsely activate only a few task-corresponding experts during training and inference.~\citet{liang2022m} propose M$^{3}$ViT, which sparsely chooses the experts by using the transformer's token embeddings for router guidance. This helps the router to assign features to a selected expert during training and inference and to reduce computational costs. 
Our proposed MoFME is orthogonal to these MoE designs
With the same goal of saving computational cost, our method instead proposes MoFME to substitute the over-parameterized parallel FFN experts with a lightweight feature modulation module followed by a single shared FFN expert. 

\noindent\textbf{Efficient MoE.}
Though MoE shows advantages in many popular tasks, its conventional architectures cannot meet requirements for practical real-world applications due to large model sizes. With many repetitive structures, pruning is the most common way to increase parameter efficiency.~\citet{wang2020deep, yang2019condconv, chen2022task} formulate channels and kernels as experts and introduce the task-specific gating network to filter out some parameters for each individual task. Several recent works~\cite{xue2022one, rajbhandari2022deepspeed} also consider applying knowledge distillation to obtain a lightweight student model for inference only. However, the above methods sacrifice model performance. Besides,~\citet{jiang2021towards, liang2022m} study how to efficiently adapt MoE networks to hardware devices while saving communication and computational costs.
Instead, our MoFME aims to decrease computational costs and targets the redundancies in conventional over-parameterized FFN experts without a drop in performance by learning lightweight feature-modulated layers.

\noindent\textbf{Adverse Weather Removal.}
Adverse weather removal has been explored in many aspects. For example, MPRNet~\cite{zamir2021multi}, SwinIR~\cite{liang2021swinir}, and Restormer~\cite{zamir2022restormer} are architectures for general image restoration. Some methods can remove multiple adverse weathers at once. All-in-One~\cite{li2020all} uses neural architecture search (NAS) to discriminate between different tasks.
TransWeather~\cite{valanarasu2022transweather} uses learnable weather-type embeddings in the decoder. Transformer is also applied in this task. UFormer~\cite{wang2022uformer} and Restormer~\cite{zamir2022restormer} construct pyramidal network structures for image restoration based on locally-enhanced windows and channel-wise self-attention, respectively.

\section{Proposed Methods}
\subsection{Feature Modulated Expert}
\label{ssec:FME}
We consider a common Mixture-of-Experts setting with the Vision Transformer (ViT) architecture~\cite{dosovitskiy2020image}, where the dense FFN in each transformer block is replaced by a Mixture-of-Experts layer.
The MoE layer inputs are $N$ tokens 
$\boldsymbol{x}\in\mathbb{R}^{D}$ from the Multi-head Attention layer. 
Each token $\boldsymbol{x}$ is assigned by an input-dependent router into a set of $E$ experts with router weight $r(\boldsymbol{x})$. 

In a typical MoE design with a linear router, the functionality of the router can be formulated as
\begin{equation}
         r(\boldsymbol{x}) = TopK(\textrm{softmax}(\textbf{W}_{r} \boldsymbol{x})),
\end{equation}
\begin{equation}
\begin{aligned}
     TopK(\text{v}) & =\left\{  
             \begin{array}{lr}  
             \text{v},\; \text{if v is in the top $K$ elements}   \\  
             0,\; \text{otherwise}    
             \end{array}  
\right.  
\end{aligned}
\end{equation}
where $\textbf{W}_{r}\in\mathbb{R}^{E\times D}$ is a trainable parameter, which maps input token into $E$ router logits for experts selection. 
To reduce the computation cost, the experts in the model are sparsely activated, with $TopK(\cdot)$ setting all elements of the router weight to zero except
the elements with the largest K values.
For clarity in the rest of the paper, we denote the router weight of the $i^{th}$ expert as $r_i(\boldsymbol{x})$.

The output of the MoE layer is therefore formulated as the weighted combination of the experts' output on the input token $\boldsymbol{x}$~\cite{shazeer2017outrageously} as
\begin{equation}
    MoE(\boldsymbol{x}) = \sum_{i} r_{i}(\boldsymbol{x})e_{i}(\boldsymbol{x}),
\end{equation}
where $e_{i}(\cdot)$ denotes a functionality of the $i^{th}$ expert, typically designed as a FFN in the context of vision transformers.
This process is illustrated in Figure~\ref{fig:method}(a).

In this work, we employ the technique of Linear Feature Modulation~\cite{perez2018film} into the design of MoE to propose the efficient Feature Modulated Expert block, as illustrated in Figure~\ref{fig:method}(b). Specifically, the diverse task-specific features, i.e. tokens, are first modulated with a dynamic feature modulation unit, where the tokens are directed to different learned affine transformations based on an input-dependent router. The modulated features are then fused by a single shared FFN expert.
In this way, we implicitly represent each expert in the MoE architecture as the cascading modules of a lightweight affine feature modulation transformation and a shared FFN, significantly reducing the parameter and computation overhead for adding additional experts.

First we formulate a single Feature Modulation (FM) block~\cite{perez2018film}. We obtain input-dependent feature modulation parameters $\gamma\in \mathbb{R}^{D}$ and $\beta\in\mathbb{R}^{D}$ with two functions $g:\mathbb{R}^{D}\rightarrow \mathbb{R}^{D}$ and $b:\mathbb{R}^{D}\rightarrow \mathbb{R}^{D}$ respectively according to an input token $\boldsymbol{x}$ as
\begin{equation}
\begin{aligned}
\gamma= g(\boldsymbol{x}) \quad\quad\quad  \beta=b(\boldsymbol{x}),
\end{aligned}
\end{equation}
where $g$ and $b$ can take arbitrary learnable functions. In practice, those functions are implemented with lightweight $1\times1$ convolutions. 
The input token is then modulated as
\begin{equation}
    FM(\boldsymbol{x}) = \gamma\circ \boldsymbol{x}+\beta,
\end{equation}
where $\circ$ is the Hadamard (element-wise) product taken w.r.t. the feature dimension.

To combine the FM module with MoE, we instantiate $E$ independent FM modules to modulate diverse task-specific features, each parameterized with  $\gamma^{(i)}$ and $\beta^{(i)}$, where $i\in \{1,...,E\}$. Adapting from the traditional MoE formulation, we let the router select which FM module to apply on the input token, rather than which FFN to be used. Specifically, our FME module is formulated as
\begin{equation}
\begin{aligned}
&FME(\boldsymbol{x}|\gamma,\beta) \\ &=  FFN \left\{ \sum_{i} r_i(\boldsymbol{x})\cdot [\gamma^{(i)}\circ \boldsymbol{x}+\beta^{(i)}] \right\},
\end{aligned}
\label{fme}
\end{equation}
where a single shared FFN module can process the mixture of multi-task features by the diverse feature modulations.

\subsection{Uncertainty-aware Router}
\label{ssec:UaR}
To improve the FME performance, we propose Uncertainty-aware Router (UaR), which performs implicit uncertainty estimation on the router weights according to MC dropout~\cite{gal2016dropout}.
Model uncertainty~\cite{lakshminarayanan2017simple} measures if the model \textit{knows what it knows}. Although there exists ensemble-based uncertainty estimation methods~\cite{ovadia2019can, ashukha2020pitfalls} that often achieve the best calibration
and predictive accuracy, the high computational complexity and storage cost motivates us to use the more efficient MC dropout~\cite{rizve2021defense}.

Specifically, we can regard the output of a certain router $r(\boldsymbol{x})$ as a Gaussian distribution to calibrate its uncertainty. The mean and covariance of such distribution can be estimated via a ``router ensemble'', where we pass the token representation $\boldsymbol{x}$ to get $r(\boldsymbol{x)}$ with the router for $M$ times according to MC dropout. We denote the resulted ensemble as $r^{m}(\boldsymbol{x})=\{r^{1}(\boldsymbol{x}),r^{2}(\boldsymbol{x}),...,r^{M}(\boldsymbol{x})\}$, and the mean and covariance of the router weights in the ensemble as $\Check{\mu}$ and $\Check{\Sigma}$ respectively. 
We calibrate and normalize the router's logits according to~\citet{al2020federated} as
\begin{equation}
\begin{aligned}
\Check{r}(\boldsymbol{x}) = \Check{\Sigma}^{-1} [ r(\boldsymbol{x})-\Check{\mu}] / || \Check{\Sigma}^{-1} [ r(\boldsymbol{x})-\Check{\mu}]||_2,
\end{aligned}
\end{equation}
where $\Check{r}(\boldsymbol{x})$ is used in the forward and backward pass during the training. The mean $\Check{\mu}$ and inverse covariance $\Check{\Sigma}^{-1}$ are both formulated as zero-padded diagonal matrices in the computation. The detailed structure is shown in Figure \ref{fig:method}.

\subsection{Optimization Objective}
MoE-based model would suffer from performance degradation if most inputs are assigned to only a small subset of experts~\cite{fedus2021switch, lepikhin2020gshard}. A load balance loss $\mathcal{L}_{lb}$~\cite{lepikhin2020gshard} is therefore proposed for MoE to penalize the number of inputs dispatched to each router:
\begin{equation}
\begin{aligned}
\label{equ:Llb}
   \mathcal{L}_{lb} = \frac{E}{N} \sum_{n=1}^{N}\sum_{i=1}^{E}v_{i}(\boldsymbol{x}_{n}) r_{i}(\boldsymbol{x}_n),
\end{aligned}
\end{equation}
where $x_n$ is the $n$-th input token, and $v_i(\boldsymbol{x}_{n})$ is 1 if the $i$-th expert is selected for $\boldsymbol{x}_{n}$ by the top-$k$ function, otherwise 0. The combined MoE training loss therefore becomes
\begin{equation}
\begin{aligned}
   \mathcal{L}_{MoE} = \mathcal{L}_{ts} + \lambda_{1} \mathcal{L}_{lb},
\end{aligned}
\end{equation}
where $\lambda_{1}$ is empirically set to $1e^{-2}$ and $\mathcal{L}_{ts}$ indicates the task-specific loss computed by model outputs and corresponding labels, e.g., MSE loss for image restoration task.

Following~\citet{lepikhin2020gshard}, we further leverage the covariance $\Check{\Sigma}$ of $r^{m}(\boldsymbol{x})$ to penalize the updating of UaR and MoFME and formulate the uncertainty loss $\mathcal{L}_{uc}$ as
\begin{equation}
\begin{aligned}
   \mathcal{L}_{uc} = \frac{E}{N} \sum_{n=1}^{N}\sum_{i=1}^{E}\Check{\Sigma}_{i} \cdot v_{i}(\boldsymbol{x}_{n}),
\end{aligned}
\end{equation}
where $v$ is defined the same as in Equation~(\ref{equ:Llb}). $\mathcal{L}_{uc}$ can further reduce the model uncertainty when optimized together with other losses, where the final MoFME objective is
\begin{equation}
\begin{aligned}
   \mathcal{L}_{MoFME} = \mathcal{L}_{ts} + \lambda_{1} \mathcal{L}_{lb} + \lambda_{2} \mathcal{L}_{uc},
\end{aligned}
\end{equation}
where $\lambda_{2}$ is empirically set to $5e^{-3}$.

\setlength\tabcolsep{6.5pt}%
\begin{table*}[t]
	\centering
 \footnotesize
 	\caption{Ablation study on All-Weather using PSNR and SSIM metrics. We set 16 experts and top2 gate.}
		\begin{tabular}{ccccccccccccc}
   \toprule
   \multirow{2}{*}{\begin{tabular}[c]{@{}c@{}}Base model\end{tabular}} & \multicolumn{2}{c}{MoFME} &  Param. &  FLOPs   & \multicolumn{2}{c}{Derain} & \multicolumn{2}{c}{Deraindrop}  & \multicolumn{2}{c}{Desnow} & \multicolumn{2}{c}{Average}\\
 \cmidrule(r){2-3} \cmidrule(r){4-4} \cmidrule(r){5-5} \cmidrule(r){6-7} \cmidrule(r){8-9}\cmidrule(r){10-11} \cmidrule(r){12-13}  
     &  FME & UaR & (M) & (GMAC) &  PSNR & SSIM & PSNR & SSIM& PSNR & SSIM& PSNR & SSIM     \\
 \cmidrule(r){4-4} \cmidrule(r){5-5} \cmidrule(r){6-6} \cmidrule(r){7-7} \cmidrule(r){8-8} \cmidrule(r){9-9} \cmidrule(r){10-10} \cmidrule(r){11-11} \cmidrule(r){12-12} \cmidrule(r){13-13}
 \multirow{1}{*}{\begin{tabular}[c]{@{}c@{}}Baseline\end{tabular}} & - & - & 8.71 & 34.93 & 27.64 & 0.9329 & 28.21 & 0.9249 & 28.40 & 0.8860 & 28.08 & 0.9146 \\
     \midrule
\multirow{4}{*}{\begin{tabular}[c]{@{}c@{}}MoE\end{tabular}} 
    & - & - & 44.19 & 37.06 & 27.91 & 0.9359 & 28.54 & 0.9307 & 28.76 & 0.8926 & 28.40 & 0.9197 \\
    & \Checkmark & - & 18.53 & 36.26 & 27.87 & 0.9342 & 28.43 & 0.9290 & 28.65 & 0.8901 & 28.32 & 0.9178 \\
    & - & \Checkmark & 44.19 & 37.17 & 27.96 & 0.9363 & 28.52 & 0.9304 & 28.80 & 0.8930 & 28.43 & 0.9199 \\
  & \Checkmark & \Checkmark & 18.53 & 36.37 & 28.01 & 0.9368 & 28.55 & 0.9311 & 28.78 & 0.8925 & 28.45 & 0.9201  \\
   \midrule
 \multirow{2}{*}{\begin{tabular}[c]{@{}c@{}}M$^{3}$ViT\end{tabular}} & - & - & 44.22 & 37.06 & 27.67 & 0.9356 & 28.42 & 0.9280 & 28.61 & 0.8911 & 28.23 & 0.9182 \\
  & \Checkmark & \Checkmark & 18.56 & 36.37 & 27.87 & 0.9344 & 28.51 & 0.9301 & 28.70 & 0.8912 & 28.36 & 0.9185\\
     \midrule
 \multirow{2}{*}{\begin{tabular}[c]{@{}c@{}}MoWE\end{tabular}} & - & - & 34.15 & 59.99 & 28.05 & 0.9370 & 28.93 & 0.9333 & 28.75 & 0.8923 & 28.58 & 0.9209 \\
  & \Checkmark & \Checkmark & 21.22 & 48.36 & 28.10 & 0.9376 & 29.03 & 0.9346 & 28.84 & 0.8927 & 28.66 & 0.9216 \\
\bottomrule
		\end{tabular}
	\label{ablation}
\end{table*}

\section{Experiments}
We evaluate our \textbf{MoFME} against several recent methods on the adverse weather removal task. We presume a test-time setup, where a model shall remove multiple types of weather effects with the same parameters. In addition, we further demonstrate the applicability of our upstream processing to downstream segmentation and classification tasks. Ablation study of MoFME architecture shows the contribution of each component. In total, MoFME achieves up to 0.1-0.2 dB performance improvement in PSNR, while saving more than 72\% of parameters and 39\% of inference time.

\subsection{Experimental Setup}
\textbf{Implementation details.}
We implement our method with the PyTorch framework using 4$\times$NVIDIA A100 GPUs. We train the network for 200 epochs with a batch size of 64. The initial learning rate of the AdamW optimizer and Cosine LR scheduler is set to $0.5\times10^{-4}$ and is gradually reduced to $10^{-6}$. We use a warm-up stage with three epochs. Input images are randomly cropped to 256$\times$256 size for training, and non-overlap crops of the same size are used at test time. We randomly flip and rotate images for data augmentation. The scaling factor for traditional MoE model is set to 4.

\setlength\tabcolsep{2pt}%
\begin{table}[t]
	\centering
 	\caption{Comparison with different numbers of experts on All-Weather. We set top2 gate in the experiments.}
	\resizebox{1\columnwidth}{!}{%
		\begin{tabular}{ccccccc}
  \toprule
            \multirow{2}{*}{} &  \multirow{2}{*}{} & 
            \# Experts   & PSNR  & FLOPs (G) & Param. (M) & Infer. time (s)     \\
            \cmidrule(r){3-3}\cmidrule(r){4-4}\cmidrule(r){5-5}\cmidrule(r){6-6}\cmidrule(r){7-7}
			\multicolumn{2}{c}{\multirow{2}{*}{\begin{tabular}[c]{@{}c@{}}MoE\\ 
        \textbf{MoFME}\end{tabular}}}     
           & 64 & 28.45 & 41.31 & 157.7 & 0.039 \\
            \multicolumn{2}{c}{\multirow{3}{*}{}}   & 64 & \textbf{28.46} & \textbf{37.43} & \textbf{47.1 (70\%$\downarrow$)}  & \textbf{0.027 (31\%$\downarrow$}) \\
            \midrule
            \midrule
			\multicolumn{2}{c}{\multirow{2}{*}{\begin{tabular}[c]{@{}c@{}}MoE\\ 
        \textbf{MoFME}\end{tabular}}}     
           & 128 & 28.56 & 41.31 & 309.1 & 0.075 \\
            \multicolumn{2}{c}{\multirow{3}{*}{}}   & 128 & \textbf{28.59} & \textbf{37.43} & \textbf{85.2 (72.5\%$\downarrow$}) & \textbf{0.046 (39\%$\downarrow$}) \\
\bottomrule
		\end{tabular}
	}
	\label{size}
\vspace{-0.5cm}
\end{table}

\setlength\tabcolsep{9pt}
\begin{table*}[h]
	\centering
 \footnotesize
 	\caption{Quantitative Comparison on Rain/HazeCityscapes using PSNR and SSIM. We set 16 experts and top4 gate.}
		\begin{tabular}{ccccccccccc}
  \toprule
            \multirow{2}*{Type} & \multirow{2}*{Method}     & \multicolumn{2}{c}{Derain}  & \multicolumn{2}{c}{Dehaze} & \multicolumn{2}{c}{Average} & Param. & FLOPs\\
                                    \cmidrule(r){3-4} \cmidrule(r){5-6} \cmidrule(r){7-8} \cmidrule(r){9-9} \cmidrule(r){10-10} 
            \multirow{2}{*}{} & \multirow{2}{*}{} & 
            PSNR$\uparrow$ & SSIM$\uparrow$  & PSNR$\uparrow$ & SSIM$\uparrow$ & PSNR$\uparrow$ & SSIM$\uparrow$   & (M)$\downarrow$  &  (GMAC)$\downarrow$  \\
                        \cmidrule(r){3-3} \cmidrule(r){4-4} \cmidrule(r){5-5} \cmidrule(r){6-6} \cmidrule(r){7-7} \cmidrule(r){8-8} \cmidrule(r){9-9} \cmidrule(r){10-10} 
            
            		\multirow{3}{*}{\begin{tabular}[c]{@{}c@{}}Task-specific\end{tabular}}  &
              RESCAN  & 19.11 & 0.9118 & 16.96 & 0.9033 & 18.04 & 0.9076 &  0.15 & 32.32 \\
            & PReNet  & 19.95 & 0.8822 & 18.22 & 0.8729 & 19.09 & 0.8776 &  0.17 & 66.58 \\
            \multirow{2}{*}{}   & FFA-Net  & 28.29 & 0.9411 & 28.96 & 0.9432 & 28.63 & 0.9422 & 4.46  & 288.34 \\
            \cmidrule(r){2-2} 
			\multirow{2}{*}{\begin{tabular}[c]{@{}c@{}}Multi-task\end{tabular}}    
             & Transweather   & 24.08 & 0.8481 & 22.56 & 0.8736 & 23.32 & 0.8609 &  38.05 & 6.12 \\
           \multirow{4}{*}{}   & Restormer   & 28.06 & 0.9630 & 22.72 & 0.9167 & 28.11 & 0.9336 & 26.13  & 140.99 \\
\midrule
			\multirow{3}{*}{\begin{tabular}[c]{@{}c@{}}Multi-task\\ MoE\end{tabular}}    
             & MoE-Vit & 32.70 & 0.9725 & 31.07 & 0.9623 & 31.89 & 0.9674 & 44.19  & 41.31 \\
            \multirow{3}{*}{}   & MMoE-Vit  & 32.47 & 0.9698 & 31.08 & 0.9582 & 31.78 & 0.9640 & 44.63  & 42.56  \\
            \multirow{3}{*}{}  & M$^{3}$ViT   & 32.56 & 0.9712 & 31.11 & 0.9597 & 31.84 & 0.9655 & 44.25  & 41.60 \\
            \multirow{3}{*}{}  & MoWE &
            \textbf{32.99} & \textbf{0.9755} & \textit{31.31} & \textit{0.9647} & \textbf{32.15} & \textbf{0.9701}  & 34.15  & 59.99 \\
\cmidrule(r){2-2} 
			\multirow{3}{*}{\begin{tabular}[c]{@{}c@{}}Efficient\\ MoE\end{tabular}}   
             & OneS  & 32.40 & 0.9691 & 30.96 & 0.9590 & 31.68 & 0.9641  &  \textbf{8.71} & \textbf{34.93} \\
            \multirow{3}{*}{}   & PR-MoE   & 32.38 & 0.9700 & 31.03 & 0.9595 & 31.71 & 0.9648  &  27.28 & 37.53 \\
                \multirow{3}{*}{}   & MoFME (ours)   & \textit{32.87} & \textit{0.9721} & \textbf{31.35} & \textbf{0.9661} & \textit{32.11} & \textit{0.9691}  & \textit{18.53} & \textit{37.43} \\
            \bottomrule
		\end{tabular}
\vspace{-0.3cm}
	\label{raincityscapes}
\end{table*}

\textbf{Metrics, datasets, and baselines.} We select widely-used PSNR and SSIM metrics as performance measures for upstream image restoration. All-weather~\cite{valanarasu2022transweather} and Rain/HazeCityscapes~\cite{hu2019depth,sakaridis2018semantic} datasets are used to evaluate deweathering and downstream segmentation. CIFAR-10 datasets is for the downstream image classification task. 

The comparison baselines include three CNN-based models RESCAN~\cite{li2018recurrent}, PRNet~\cite{ren2019progressive}, and FFA-Net~\cite{qin2020ffa} that employ task-specific weather removal. Also, we experiment with recent transformer-based models: Restormer~\cite{zamir2022restormer} with a general multi-task image restoration objective, TransWeather~\cite{valanarasu2022transweather} with learnable weather embeddings in the decoder to remove multiple adverse effects simultaneously, conventional MoE~\cite{shazeer2017outrageously}, MMoE~\cite{ma2018modeling}, M$^{3}$ViT~\cite{liang2022m}, and MoWE~\cite{luo2023mowe} for multi-task learning, as well as efficient MoE methods such as OneS~\cite{xue2022one}, which fuses the experts' weight and adopt knowledge distillation for better performance and PR-MoE~\cite{rajbhandari2022deepspeed}, which propose a pyramid residual MoE architecture to demonstrate the superiority of our proposed MoFME to handle multiple tasks in both effectiveness and efficiency. We take Vision Transformer as the backbone for MoE-based methods. 

\begin{figure}[t]
\includegraphics[width=0.45\textwidth]{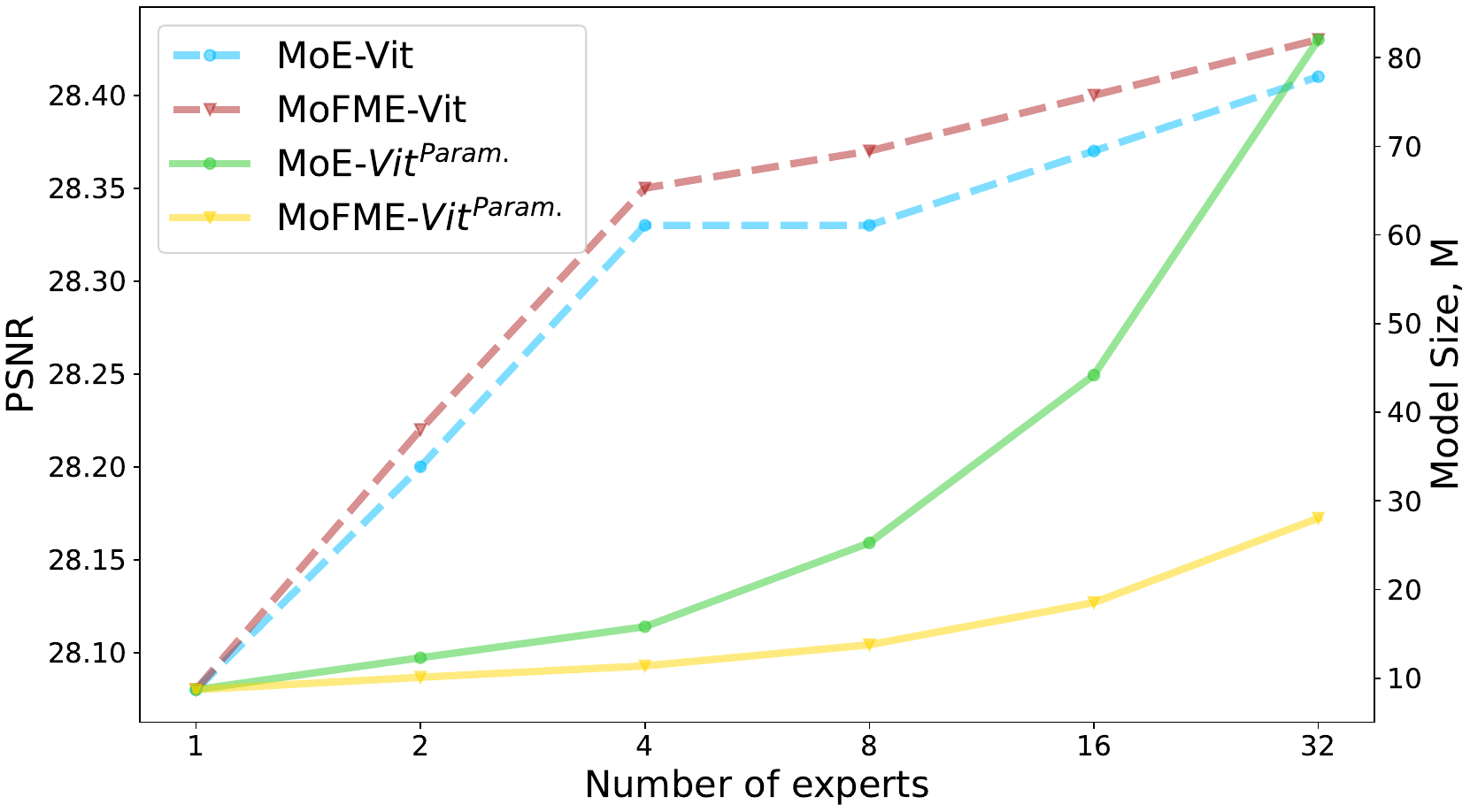}
\centering
\vspace{-0.3cm}
\caption{Number of experts v.s. PSNR and the model size. All models are trained for 100 epochs. We set top2 gate.}
\label{fig:prestudy}
\vspace{-0.5cm}
\end{figure}

\subsection{Ablation study}
We conduct ablation experiments to analyze how each proposed module contributes to the MoE performance in Table~\ref{ablation}. Starting from a traditional MoE design (base model), we replace the parallel FFN experts with FME, and examine the effectiveness of UaR by introducing the MC dropout into the router. The results suggest that FME alone can achieve significant parameter efficiency with a small performance drop, while UaR can enhance model performance by over 0.05 dB. We also apply our method on different base model including traditional MoE, M$^{3}$ViT, and MoWE, the results prove that combining the two techniques leads to improvements in both efficiency and performance for all base model.

One key property of the MoE model is its scalability while increasing the number of experts.
In Figure \ref{fig:prestudy} and Table \ref{size}, we show that the efficiency of our proposed MoFME is consistently maintained as the number of experts scales to hundreds with only 1/4-th of the parameters and over 0.1 dB improvement on All-Weather when compared to the conventional MoE. The inference time is significantly reduced by nearly 40\% when utilizing 128 experts. as shown in Table \ref{size}. 

\subsection{Quantitative analysis}

\setlength\tabcolsep{4pt}%
\begin{table}[t]
	\centering
 \footnotesize
 	\caption{Downstream semantic segmentation results after deweathering on Cityscapes using mIoU and mAcc. The expert number and topk settings are the same as Table \ref{raincityscapes}.}
		\begin{tabular}{cccccc}
  \toprule
            \multirow{2}*{Type} & \multirow{2}*{Method}     & \multicolumn{2}{c}{mIoU} & \multicolumn{2}{c}{mAcc}   \\
            \cmidrule(r){3-4} \cmidrule(r){5-6}
            \multirow{2}{*}{}  & \multirow{2}{*}{} & 
            Derain & Dehaze  & Derain & Dehaze    \\
            \cmidrule(r){2-2}\cmidrule(r){3-3} \cmidrule(r){4-4} \cmidrule(r){5-5} \cmidrule(r){6-6}
		\multirow{4}{*}{\begin{tabular}[c]{@{}c@{}}Multi-task\\ MoE\end{tabular}}    
             & MoE   & 0.4652 & 0.4541 & 0.7684 & 0.7443 \\ 
            \multirow{4}{*}{}   & MMoE  & 0.4621 & 0.4530 & 0.7643 & 0.7418 \\
            \multirow{4}{*}{}   & M$^{3}$ViT  & 0.4634 & 0.4525 & 0.7662 & 0.7421 \\
            \multirow{4}{*}{}   & MoWE & 0.4686 & 0.4545 & 0.7701 & 0.7473 \\
            \cmidrule(r){2-2}
		\multirow{3}{*}{\begin{tabular}[c]{@{}c@{}}Efficient\\ MoE \end{tabular}}     
             & OneS  & 0.4620 & 0.4519 & 0.7665 & 0.7402   \\
            \multirow{3}{*}{}   & PR-MoE  & 0.4632 & 0.4528 & 0.7660 & 0.7410 \\
            \multirow{3}{*}{}   & MoFME    & 0.4650 & 0.4550 & 0.7681 & 0.7480 \\
            \bottomrule
		\end{tabular}
\vspace{-0.5cm}
	\label{real_results}
\end{table} 

\textbf{Upstream tasks} In Table \ref{raincityscapes} and \ref{allweather}, we report the PSNR and SSIM of each type of weather and the average scores for each baseline and MoFME on All-Weather~\cite{chen2021all} and RainCityscapes~\cite{hu2019depth} after training for 200 epochs. We denote the best results in bold, and the second-best results in italics. It should be noted that all the experiments are trained with a mixture of weather data and inference with a specific type of weather. The results of Table \ref{raincityscapes} and \ref{allweather} reveal the advantage of MoE networks to deal with multi-task inputs compared with previous na\"ive transformer-based and CNN-based methods. However, as it is specifically designed for high-level tasks, M$^{3}$ViT fails to exert good performance on deweather tasks on both datasets. Furthermore, current efficient MoE methods like OneS and PR-MoE cannot achieve comparable performance compared with SOTA MoE networks, while MoFME can achieve 29.09 dB average PSNR score and 0.9272 average SSIM on All-Weather, and 32.11 dB PSNR and 0.9691 SSIM on RainCityscapes. While the MoWE model attains superior performance metrics, it is worth noting that both the model's size and its computational complexity, as quantified by the FLOPs, are substantially greater when compared to our traditional MoE-based approach.

We also provide the FLOPs and the number of parameters for each baseline on RainCityscapes in Table \ref{raincityscapes}. The MoE-based methods can achieve very satisfying scores on PSNR and SSIM. However, the heavy network structure prevents them from practical applications. The two efficient MoE baselines exert their advantages in computational costs as PR-MoE can save about 50\% parameters, and OneS merges its parameters to become a lightweight dense model. However, the certain model performance of the two methods is also sacrificed as OneS decrease almost 0.2 dB in PSNR. Our proposed MoFME takes a step forward by realizing a satisfied trade-off as it achieves compatible results compared to other SOTA baselines while saving up to 72\% parameters.

\setlength\tabcolsep{2.3pt}
\begin{table}[t]
	\centering
 \footnotesize
 	\caption{Top-1 accuracy of image classification tasks. We set the number of experts 8 and the top2 gate.}
		\begin{tabular}{ccccccc}
   \toprule
           Methods & MoE  & Efficient & Param.(M) & FLOPs(G)  & CIFAR-10      \\
            \cmidrule(r){2-2} \cmidrule(r){3-3} \cmidrule(r){4-4} \cmidrule(r){5-5} \cmidrule(r){6-6} \cmidrule(r){7-7}
            ViT
           & - & - & 13.06 & 0.85 & 98.21\% \\
            MoE-ViT 
           & \Checkmark & - & 46.16 & 1.03 & 98.33\% \\
            OneS
           & \Checkmark & \Checkmark & 13.06 & 0.85 & 98.14\% \\
            MoFME-ViT     
           & \Checkmark &  \Checkmark & 18.05 & 0.94 & 98.47\% \\
            \bottomrule
		\end{tabular}
\vspace{-0.5cm}
	\label{classiciation}
\end{table} 

\setlength\tabcolsep{8pt}%
\begin{table*}[h]
	\centering
 \footnotesize
 	\caption{Quantitative comparison on All-Weather using PSNR and SSIM metrics. We set 16 experts and top2 gate. }
  \vspace{-0.3cm}
		\begin{tabular}{cccccccccc}
   \toprule
\multirow{2}*{Type} & \multirow{2}*{Method}     & \multicolumn{2}{c}{Derain} & \multicolumn{2}{c}{Deraindrop}  & \multicolumn{2}{c}{Desnow} & \multicolumn{2}{c}{Average}\\
            \cmidrule(r){3-4} \cmidrule(r){5-6} \cmidrule(r){7-8} \cmidrule(r){9-10}
\multirow{2}{*}{}  & \multirow{2}{*}{} & 
            PSNR$\uparrow$ & SSIM$\uparrow$  & PSNR$\uparrow$ & SSIM$\uparrow$ & PSNR$\uparrow$ & SSIM$\uparrow$ & PSNR$\uparrow$ & SSIM$\uparrow$      \\
                                    \cmidrule(r){3-3} \cmidrule(r){4-4} \cmidrule(r){5-5} \cmidrule(r){6-6} \cmidrule(r){7-7} \cmidrule(r){8-8} \cmidrule(r){9-9} \cmidrule(r){10-10} 
			\multirow{3}{*}{\begin{tabular}[c]{@{}c@{}}Task-specific\end{tabular}} &
             RESCAN  & 21.57 &  0.7255 & 24.26 & 0.8367 & 24.30 & 0.7586 & 23.38 & 0.7736 \\
             & PReNet  & 23.16 & 0.8624 & 24.96 & 0.8629 & 25.19 & 0.8483 & 24.44 & 0.8579 \\
            \multirow{2}{*}{}   & FFA-Net  & 27.96 & 0.8857 & 27.73 & 0.8894 & 27.21 & 0.8578 & 27.63 & 0.8776 \\
\cmidrule(r){2-2}
			\multirow{2}{*}{\begin{tabular}[c]{@{}c@{}}Multi-task\end{tabular}}    
             & Transweather   & 25.64 & 0.8103 & 27.37 & 0.8570 & 26.98 & 0.8305 & 26.66 & 0.8326 \\
            \multirow{4}{*}{}   & Restormer   & 27.85 & 0.8802 & 28.32 & 0.8881 & 28.18 & 0.8684 & 28.12 & 0.8789 \\
\midrule
		\multirow{4}{*}{\begin{tabular}[c]{@{}c@{}}Multi-task\\ MoE\end{tabular}}     
             & MoE-Vit  & 
             28.47 & 0.9420 & 29.06 & 0.9367 & 29.20 & 0.8987 & 28.91 & 0.9258 \\
            \multirow{4}{*}{}  & MMoE-Vit  & 28.52 & 0.9415 & 28.91 & 0.9368 & 29.13 & 0.8986 & 28.85 & 0.9256 \\
            \multirow{4}{*}{}  & M$^{3}$ViT 
            & 28.61 & 0.9428 & 28.75 & 0.9345 & 29.27 & 0.9004 & 28.88 & 0.9259 \\
            \multirow{4}{*}{}  & MoWE & 
            \textit{28.59} & \textit{0.9432} & \textbf{29.37} & \textbf{0.9400} & \textbf{29.37} & \textbf{0.9014} & \textbf{29.11} & \textbf{0.9282}\\
\cmidrule(r){2-2}
			\multirow{3}{*}{\begin{tabular}[c]{@{}c@{}}Efficient\\ MoE\end{tabular}}     
             & OneS    & 28.35 & 0.9384 & 28.89 & 0.9341 & 28.98 & 0.8976 & 28.74 & 0.9234 \\
            \multirow{3}{*}{}   & PR-MoE   & 28.43 & 0.9394 & 28.97 & 0.9342 & 29.18 & 0.8980 & 28.86 & 0.9239 \\
            \multirow{3}{*}{}   & MoFME(ours)   & \textbf{28.66} & \textbf{0.9436} & \textit{29.27} & \textit{0.9385} & \textit{29.35} & \textit{0.8996} & \textit{29.09} & \textit{0.9272}\\

            \bottomrule
		\end{tabular}
\vspace{-0.3cm}
	\label{allweather}
\end{table*} 

\begin{figure*}[t]
\includegraphics[width=0.9\textwidth]{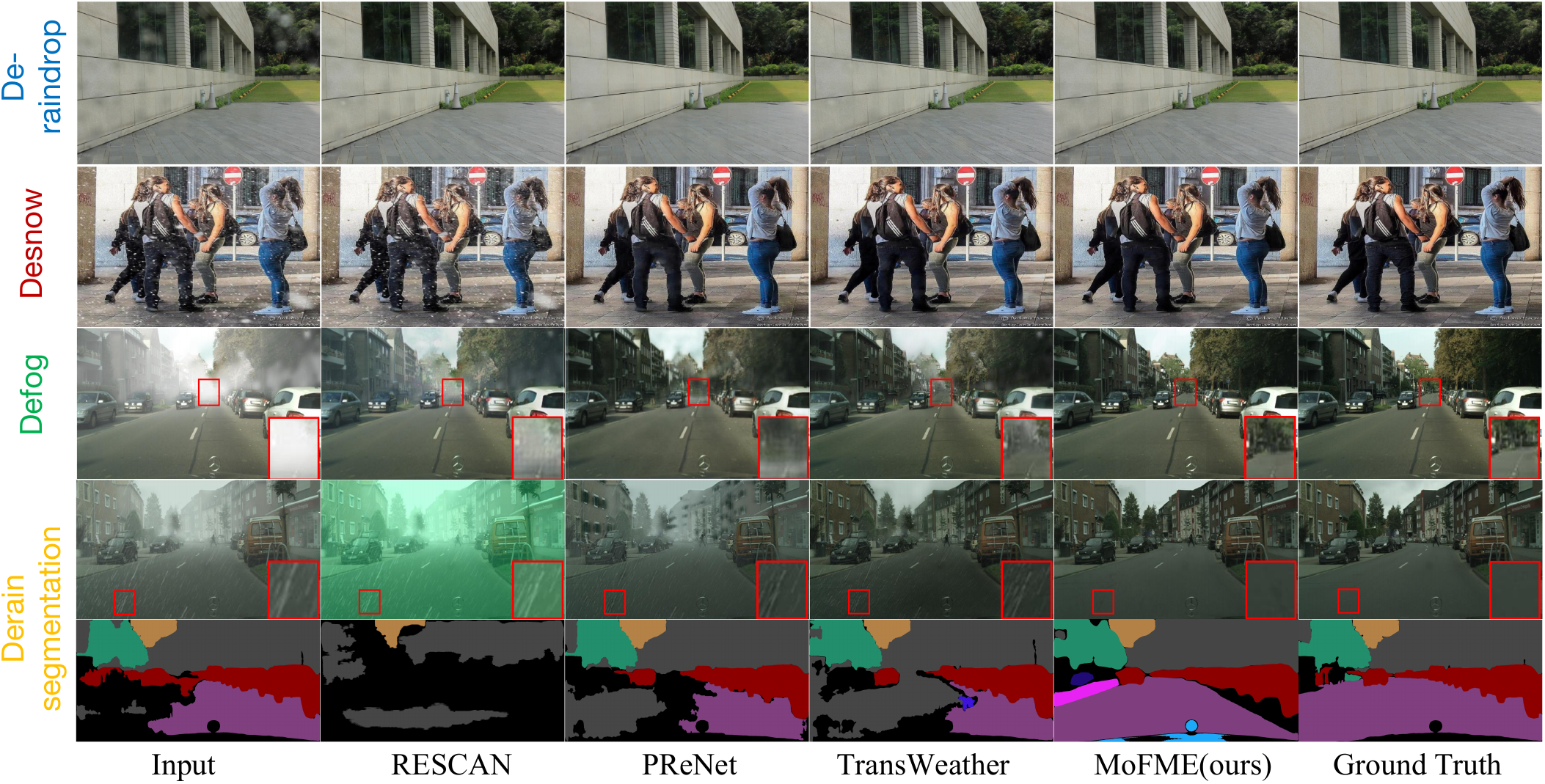}
\centering
\vspace{-0.12cm}
\caption{Examples of noisy inputs (left), noise-free ground-truth (right), and the methods' deweathering results. We show the upstream image restoration (top) and the effects on the downstream segmentation task (bottom) for RainCityscapes.}
\label{fig:vis_up}
\end{figure*}

\textbf{Downstream task}
\ding{202} \textit{Semantic segmentation:} Although our proposed methods exert satisfying performance with efficiency on low-level image restoration tasks, however, as has been questioned by~\citet{liu2022exploring}, \textit{will images optimized for better human perception can be accurately recognized by machines?} 
We provide the quantitative comparison on Cityscapes for downstream segmentation tasks based on mIoU and mAcc in Table \ref{real_results}. We can find that other efficient MoE baselines fail to make satisfying predictions on the downstream task.  On the other hand, our proposed MoFME exerts satisfying performance on both the upstream deweather task and downstream task by outperforming 2\% mIoU and 2.5\% mAcc compared with other efficient MoE baselines. We also provide the visualization results in Figure \ref{fig:vis_up}.
\ding{203} \textit{Image classification:} To further prove the generality of our methods, we perform image classification task on CIFAR-10 with ImageNet pre-training. The top-1 accuracy is reported in Table~\ref{classiciation} which shows that MoE models lead to performance gain with parameter costs, while MoFME outperforms other similar size baselines by 0.2\% on CIFAR-10.

\subsection{Qualitative analysis}
Visual results in Figure \ref{fig:vis_up} show the qualitative comparison of our method against the other methods. As shown in the top three rows, MoFME can achieve better visual results compared with previous methods, which recovers sharper information of the original image, especially in the defog setting. The visual results also demonstrate that our method can further recover downstream task-friendly images with better semantic segmentation outcomes. Our proposed MoFME is able to segment out clearer boundaries while maintaining consistency in color and texture.

\section{Conclusion}
In this work, we proposed Mixutre-of-Feature-Modulation-Experts (MoFME) approach with novel Feature Modulation Expert (FME) and Uncertainty-aware Router (UaR). 
Extensive experiments on deweathering task demonstrated that MoFME can handle multiple tasks simultaneously, as it outperformed prior MoE-based baselines by 0.1-0.2 dB while saving more than 72\% of parameters and 39\% inference time. Downstream classification and segmentation results proved MoFME generalization to real-world applications.

\clearpage
\section{Acknowledgments}
Shanghang Zhang is supported by the National Key Research and Development Project of China (No.2022ZD0117801). The authors would like to express their sincere gratitude to the Interdisciplinary Research Center for Future Intelligent Chips (Chip-X) and Yachen Foundation  for their invaluable support. 
\bigskip

\bibliography{aaai24}

\end{document}


\maketitle

In the supplementary material, we first present more details of the deweather datasets in Section 1. We then present our modified Vision Transformer with a shared head and tail to better coordinate with the deweather task in Section 2. In Section 3, we provide additional visualization results. Then, we provide an alternative efficient solution for UaR with compatible performance in Section 4. Finally, we provide the complete version of Table \ref{ap:real_results} due to the page limitation.

\section{Detailed metrics and datasets}
\label{sec:1}
We adopt Peak Signal-to-Noise Ratio (PSNR) and Structural Similarity (SSIM) as the quantitative evaluation metrics while using weather-free images as ground truths. As for further downstream semantic segmentation tasks with RainCityscapes~\cite{hu2019depth}, we adopt mIoU and mAcc as the quantitative evaluation metric. For a fair comparison, all experiments are trained on the mix of all types of weather and tested on specific types of weather. 

All-weather~\cite{valanarasu2022transweather} collects data from different public datasets. The training set is composed of 8250 rain streaks and haze images from Outdoor-Rain~\cite{li2019outdoor-rain}, 861 raindrop images from Raindrop~\cite{qian2018raindrop}, and 8958 snow images from Snow100K~\cite{liu2018snow100k}. The test set is made up of Test1 dataset which is sampled from Outdoor-Rain~\cite{li2019outdoor-rain}, the Raindrop test set~\cite{qian2018raindrop}, and the Snow 100k-L test set~\cite{liu2018snow100k}. RainCityscapes~\cite{hu2019depth} is based on Cityscapes~\cite{cordts2016cityscapes} dataset with additional synthetic rain streaks and fog apply to ground-truth rain-free photos.

\section{Supplementary description of vision transformer with shared head and tail}
\label{sec:2}

To better coordinate with the deweather task, we modify the vision transformer~\cite{dosovitskiy2020image} to have a task-shared head and tail and take it as our baseline. The motivation comes from the fact that Transformer exerts great advantages in modeling long-distance relationships and dynamic processing.

Specifically, we only adopt a single task-shared head to obtain image low-level features since we do not know the weather type of the inputs in advance. Given an image $\textbf{I} \in \mathbb{R}^{3\times H \times W}$, we denote the image feature $\boldsymbol{x} \in \mathbb{R}^{C \times H \times W}$ with channel $C$ as $ \boldsymbol{x} = Head(\textbf{I})$, where $Head(\cdot)$ suggests the task-shared head. For simplicity, we adopt the same head setting as IPT\cite{chen2021ipt}, which consists of one convolution layer and two residual blocks. The former parameters are $3 \times 3$ kernel size, $3$ input channels, and $32$ output channels. The latter consists of two convolution layers with $3 \times 3$ kernel size, $32$ input channels, and $32$ output channels, which involve a shortcut. The size of feature maps remains unchanged in the process.

The image feature $\boldsymbol{x} \in \mathbb{R}^{C \times H \times W}$ will be then divided into $N=\frac{HW}{P^2}$ non-overlap patches $x_i \in \mathbb{R}^{C \times \frac{H}{P} \times \frac{W}{P}}, i=\{ 1, \dots N \}$, where $P$ indicates the size of the patch. Then the 2D convolution $\textbf{C}$ is utilized to project the patches $\{x_i\}$ into low-dimension embedding $\{ t_i = \textbf{C}x_i \in \mathbb{R} ^ {D} \}$. We further apply the learnable position embedding $\textbf{PE} \in \mathbb{R}^{N\times D}$ to preserve the spatial relationship of the patch embeddings. Finally, we formulate the input tokens of vision transformer $\textbf{t} \in \mathbb{R}^{N\times D}$ as:

\begin{equation}
    \textbf{t} = [\textbf{C}x_1, \textbf{C}x_2, \dots, \textbf{C}x_N] + \textbf{PE}
\end{equation}

The tokens $\textbf{t}$ will be then fed into the transformer encoder consisting of $L$ transformer blocks with spatial multi-headed self-attention (S-MHSA), layer normalization (LN) and feed-forward network (FFN), which can be represented as:
\begin{align}
    \label{self-attention}
    \textbf{y}^{\ell} & = \text{S-MHSA}(\text{LN}(\textbf{t}^{\ell})) + \textbf{t}^{\ell} \\
    \textbf{t}^{\ell +1} & = \text{FFN}(\text{LN}(\textbf{y}^{\ell})) + \textbf{y}^{\ell}    
\end{align}
where $\textbf{y}^{\ell}$ is the output of S-MHSA in $\ell=\{1, \dots, L\}$ layer of the Transformer blocks, and FFN is made up of two fully-connected layers with GELU and dropout.

\label{sec:3}
\begin{figure}[t]
\centering
\subfigure[OneS]{\includegraphics[width=4cm]{AnonymousSubmission/LaTeX/images/6_vis.png}}
\subfigure[MoE]{\includegraphics[width=4cm]{AnonymousSubmission/LaTeX/images/1_vis.png}}
\subfigure[MoWE]{\includegraphics[width=4cm]{AnonymousSubmission/LaTeX/images/5_vis.png}}
\subfigure[MoFME]{\includegraphics[width=4cm]{AnonymousSubmission/LaTeX/images/2_vis.png}}
\vspace{-0.4cm}
\caption{T-SNE visualization of the output embeddings from the MoE layers with five experts.} 
\vspace{-0.6cm}
\label{fig:tsne_app}  
\end{figure}

\begin{figure*}[t]
\includegraphics[width=\textwidth]{AnonymousSubmission/LaTeX/images/vis_appendix.pdf}
\centering
\caption{Qualitative results to visualize the deweather performance. We provide the RGB images of model input and predictions. The deweather types from top to bottom are \textcolor{red}{desnow}, \textcolor{blue}{deraindrop}, and \textcolor{orange}{derain} on All-Weather, and \textcolor{green}{defog} with downstream segmentation on RainCityscapes.}
\label{fig:vis_a}
\end{figure*}

We then apply a \textbf{Inverted Bottleneck} to inverse the patch embedding to restore the origin resolution instead of common up-sampling methods like pixel-shuffle~\cite{shi2016real} or interpolation~\cite{chen2021pre, valanarasu2022transweather}. The inverted bottleneck is a linear layer that transforms the dimension of $\textbf{t}^{L} \in \mathbb{R}^{N \times D} = \textbf{t}^{L} \in \mathbb{R}^{(\frac{H}{P} \times \frac{W}{P}) \times D}$ to $\textbf{t}^{L} \in \mathbb{R}^{(\frac{H}{P} \times \frac{W}{P}) \times (C \times P \times P)}$. Thus, $\textbf{t}^{L}$ can be further reshaped into $\textbf{t}^{L} \in \mathbb{R}^{C \times H \times W}$ which shares the same dimension as input image feature $\boldsymbol{x} \in \mathbb{R}^{C \times H \times W}$. The benefit of the inverted bottleneck is mainly due to the large number of learnable parameters, which preserves the spatial relations of the patch embeddings.

Finally, the task-shared tail consists of $4$ sequential blocks to adjust the channels of feature $\textbf{t}^{L}$ to obtain the restoration clean image $\textbf{I}_{clean} \in \mathbb{R}^{3\times H \times W}$ according to $\textbf{I}_{clean} = tail(\textbf{t}^{L})$. The residual block and convolution layer are used twice alternately. The input channels are $256$ and the output channels of each layer are $256, 128, 128, 3$ respectively. $3 \times 3$ kernel size is adopted for all of them.

\setlength\tabcolsep{4pt}
\begin{table}[t]
	\centering
 \footnotesize
 	\caption{Downstream semantic segmentation results after deweathering on Cityscapes using mIoU and mAcc.}
		\begin{tabular}{cccccc}
  \toprule
            \multirow{2}*{Type} & \multirow{2}*{Method}     & \multicolumn{2}{c}{mIoU} & \multicolumn{2}{c}{mAcc}   \\
            \cmidrule(r){3-4} \cmidrule(r){5-6}
            \multirow{2}{*}{}  & \multirow{2}{*}{} & 
            Derain & Dehaze  & Derain & Dehaze    \\
            \cmidrule(r){2-2}\cmidrule(r){3-3} \cmidrule(r){4-4} \cmidrule(r){5-5} \cmidrule(r){6-6}
		\multirow{2}{*}{\begin{tabular}[c]{@{}c@{}}Task\\ Specific\end{tabular}}   
             & PReNet     &  0.1321 & 0.3305 & 0.2943 & 0.2508  \\
            \multirow{4}{*}{}   & MSBDN-DFF   &  0.1744 & 0.3426 &  0.2733 & 0.4019 \\
            \cmidrule(r){2-2}
			\multirow{2}{*}{\begin{tabular}[c]{@{}c@{}}Multi\\ Task\end{tabular}}    
             & Transweather     &  0.4425 & 0.3643 &  0.6710 & 0.6105  \\
            \multirow{4}{*}{}   & Restormer   &  0.4383 & 0.4085 &  0.6833 & 0.6920 \\
\midrule
		\multirow{4}{*}{\begin{tabular}[c]{@{}c@{}}Multi-task\\ MoE\end{tabular}}    
             & MoE   & 0.4532 & 0.4494 & 0.7382 & 0.7240 \\ 
            \multirow{4}{*}{}   & MMoE  & 0.4585 & 0.4521 & 0.7462 & 0.7392 \\
            \multirow{4}{*}{}   & M$^{3}$ViT  & 0.4556 & 0.4539 & 0.7396 & 0.7305 \\
            \multirow{4}{*}{}   & MoWE & 0.4686 & 0.4545 & 0.7701 & 0.7473 \\
            \cmidrule(r){2-2}
		\multirow{3}{*}{\begin{tabular}[c]{@{}c@{}}Effcient\\ MoE \end{tabular}}     
             & PR-MoE    & 0.4379 & 0.4367 & 0.7245 & 0.7148   \\
            \multirow{3}{*}{}   & OneS    & 0.4290 & 0.4310 & 0.7196 & 0.7101 \\
            \multirow{3}{*}{}   & MoFME    & 0.4678 & 0.4548 & 0.7754 & 0.7580 \\
            \bottomrule
		\end{tabular}
\vspace{-0.5cm}
	\label{ap:real_results}
\end{table} 

\section{Additional visualization results}
We visualize the output of the different kinds of MoE layers. We can clearly see that the traditional MoE can not distinguish the diverse input clearly, while the efficient MoE method OneS~\cite{xue2022one} fails to tell the difference between different types of embeddings. Our proposed MoFME, on the other hand, can correlate the features well with clear areas and boundaries.

We further illustrate more visualized predictions from All-Weather and Raincityscapes in Figure \ref{fig:vis_a}. The deweather types from top to bottom are \textcolor{red}{Desnow}, \textcolor{blue}{Deraindrop}, and \textcolor{orange}{Derain} on All-Weather, and \textcolor{green}{Defog} with downstream segmentation on RainCityscapes. Compared to the other baselines, it can be seen that our method achieves visually pleasing results on different deweather tasks. Take visualization in All-Weather as an example, our method works better in removing a mixture of raindrops, snowflakes, and fog, while other methods fail to deal with the rain and snow removal and result in artifacts. 

\section{Mathematical proof and alternative uncertainty estimation method}
\label{sec:4}
In this section, we can derive a loss function based
on maximizing the Gaussian likelihood Consider the least square regression with a neural network, which is learned on dataset $\mathcal{D}:=(\boldsymbol{X},y)$ with parameter $\boldsymbol{\theta}$. The quadratic min square error objective $f(\boldsymbol{\theta})$ can be formulated as
and loss $f(\boldsymbol{\theta};\boldsymbol{x},y):=\frac{1}{2}(\boldsymbol{X}^{T}\boldsymbol{\theta}-y)^{2}$ is quadratic. The primary objective function can be represented with model parameter $\boldsymbol{\theta}\in\mathbb{R}^{d}$ as
\begin{equation}
\begin{aligned}
f(\boldsymbol{\theta}) &= \frac{1}{2}||\boldsymbol{X}\boldsymbol{\theta}-\boldsymbol{y}||^{2}
=\log\left[ e^{\frac{1}{2} ||\boldsymbol{X}\boldsymbol{\theta}-\boldsymbol{y}||^{2}} \right]\\
&=\log\left[\mathcal{N}(\boldsymbol{\theta}|\mu,\Sigma)\right]+const,
\end{aligned}
\end{equation}
where $\Sigma^{-1}:=\boldsymbol{X^{T}X}$ and $\mu:=(\boldsymbol{X^{T}X})^{-1}\boldsymbol{X^{T}y}$ are the inverse covariance and mean of a multivariate Gaussian distribution of model parameter $\theta$, and the objective formulated as the log-likelihood. Therefore, we can take the mean and covariance of the intermediate features to estimate the model uncertainty.

Besides, the calculation of co-variance is time-consuming while training sample times, especially the covariance are hard to obtain. So we also propose an alternative uncertainty estimation method to simplify this process. With a similar precondition mentioned in UaR section, we sample only the output of each training to calculate the KL divergence as an uncertainty factor to represent the efficiency of this training. Therefore, we calibrate the router's logits as a uniformly-weighted mixture distribution and combine ensemble of router's weight logits $\{r^{(1)}(\mathbf{x}),...,r^{(M)}(\mathbf{x})\}$ according to MC dropout~\cite{rizve2021defense}, where $r^{(m)}(\mathbf{x})\in \mathbb{R}^{E}$, as
\begin{equation}
\begin{aligned}
\Check{r}(\mathbf{x}) = M^{-1}\sum_{m=1}^{M}r^{(m)}(\mathbf{x})
\end{aligned}
\end{equation}
the optimization objection will remain the same for reducing the model uncertainty. Such a method simplifies the computation of UaR during the MoFME training period while sacrificing parts of the model performance. We provide such an alternative method for the piratical employment of MoFME with limited computational resources.

\section{Additional experiment results}
We here provide the complete version of Table \ref{ap:real_results} in Table~\ref{ap:real_results} due to the page limitation for better comparison and further demonstrate the superiority of our proposed MoFME method.

\bibliography{aaai24}